\documentclass[10pt,twocolumn,letterpaper]{article}

\usepackage[accsupp]{axessibility}  
\usepackage{color,xcolor}
\usepackage{epsfig}
\usepackage{graphicx}

\usepackage{adjustbox}
\usepackage{array}
\usepackage{booktabs}
\usepackage{colortbl}
\usepackage{wrapfig}
\usepackage{hhline}
\usepackage{multirow}
\usepackage{wrapfig}
\usepackage{floatflt}

\usepackage{bbm}
\usepackage{amsmath,amsfonts,amssymb}
\usepackage{mathtools}  %
\usepackage{bm}
\usepackage{nicefrac}
\usepackage{microtype}

\usepackage{changepage}
\usepackage{extramarks}
\usepackage{fancyhdr}
\usepackage{lastpage}
\usepackage{setspace}
\usepackage{soul}
\usepackage{xspace}

\usepackage{enumerate}
\usepackage{enumitem}  %

\usepackage{makecell}

\usepackage{pifont} %

\usepackage{algorithm,algpseudocode}

\usepackage{amsthm}

\usepackage[symbol]{footmisc}
\newcolumntype{L}[1]{>{\raggedright\let\newline\\\arraybackslash\hspace{0pt}}m{#1}}
\newcolumntype{C}[1]{>{\centering\let\newline\\\arraybackslash\hspace{0pt}}m{#1}}
\newcolumntype{R}[1]{>{\raggedleft\let\newline\\\arraybackslash\hspace{0pt}}m{#1}}

\newcommand{\ignore}[1]{}

\makeatletter
\DeclareRobustCommand\onedot{\futurelet\@let@token\@onedot}
\def\@onedot{\ifx\@let@token.\else.\null\fi\xspace}

\def\eg{e.g\onedot} 
\def\ie{i.e\onedot} 
 
\def\etc{etc\onedot}

\makeatother

\definecolor{MyDarkBlue}{rgb}{0,0.08,0.8}
\definecolor{MyDarkGreen}{RGB}{45,155,45}
\definecolor{MyDarkRed}{rgb}{0.8,0.02,0.02}
\definecolor{MyDarkOrange}{rgb}{0.40,0.2,0.02}
\definecolor{MyPurple}{RGB}{111,0,255}
\definecolor{MyRed}{rgb}{0.8,0.0,0.0}
\definecolor{MyGold}{rgb}{0.75,0.6,0.12}
\definecolor{MyDarkgray}{rgb}{0.66, 0.66, 0.66}
\definecolor{JiayuanColor}{rgb}{0.60,0.43,0.48}

\newcommand{\model}{NS3D\xspace}

\newcommand{\xhdr}[1]{{\noindent \textbf{#1}}}

\newcommand{\mycell}[1]{\begin{tabular}[t]{@{}l@{}l}#1\end{tabular}}

\usepackage{amsmath,amsfonts,bm}

\def\eqref#1{equation~\ref{#1}}

\def\1{\bm{1}}

\DeclareMathAlphabet{\mathsfit}{\encodingdefault}{\sfdefault}{m}{sl}
\SetMathAlphabet{\mathsfit}{bold}{\encodingdefault}{\sfdefault}{bx}{n}

\def\gE{{\mathcal{E}}}

\def\gO{{\mathcal{O}}}

\def\gQ{{\mathcal{Q}}}

\def\gU{{\mathcal{U}}}

\newcommand{\sigmoid}{\sigma}

\usepackage{inc/cvpr}              %

\usepackage[accsupp]{axessibility}  %

\usepackage{listings}
\usepackage[pagebackref,breaklinks,colorlinks]{hyperref}

\usepackage[capitalize]{cleveref}
\crefname{section}{Sec.}{Secs.}
\Crefname{section}{Section}{Sections}
\Crefname{table}{Table}{Tables}
\crefname{table}{Tab.}{Tabs.}

\def\cvprPaperID{2279} %
\def\confName{CVPR}
\def\confYear{2023}

\begin{document}

\title{\model: Neuro-Symbolic Grounding of 3D Objects and Relations}

\author{Joy Hsu\\
Stanford University\\
{\tt\small joycj@stanford.edu}
\and
Jiayuan Mao \\
Massachusetts Institute of Technology\\
{\tt\small jiayuanm@mit.edu}
\and
Jiajun Wu \\
Stanford University\\
{\tt\small jiajunwu@cs.stanford.edu}
}
\maketitle

\begin{abstract}
Grounding object properties and relations in 3D scenes is a prerequisite for a wide range of artificial intelligence tasks, such as visually grounded dialogues and embodied manipulation. However, the variability of the 3D domain induces two fundamental challenges: 1) the expense of labeling and 2) the complexity of 3D grounded language. Hence, essential desiderata for models are to be data-efficient, generalize to different data distributions and tasks with unseen semantic forms, as well as ground complex language semantics (\eg, view-point anchoring and multi-object reference). To address these challenges, we propose \model, a neuro-symbolic framework for 3D grounding. \model translates language into programs with hierarchical structures by leveraging large language-to-code models. Different functional modules in the programs are implemented as neural networks. Notably, \model extends prior neuro-symbolic visual reasoning methods by introducing functional modules that effectively reason about high-arity relations (\ie, relations among more than two objects), key in disambiguating objects in complex 3D scenes. Modular and compositional architecture enables \model to achieve state-of-the-art results on the ReferIt3D view-dependence task, a 3D referring expression comprehension benchmark. Importantly, \model shows significantly improved performance on settings of data-efficiency and generalization, and demonstrate zero-shot transfer to an unseen 3D question-answering task. %

\end{abstract}
\vspace{-1em} %
\section{Introduction}
\label{sec:intro}

\begin{figure}[tp!]
  \centering
  \includegraphics[width=1.0\linewidth]{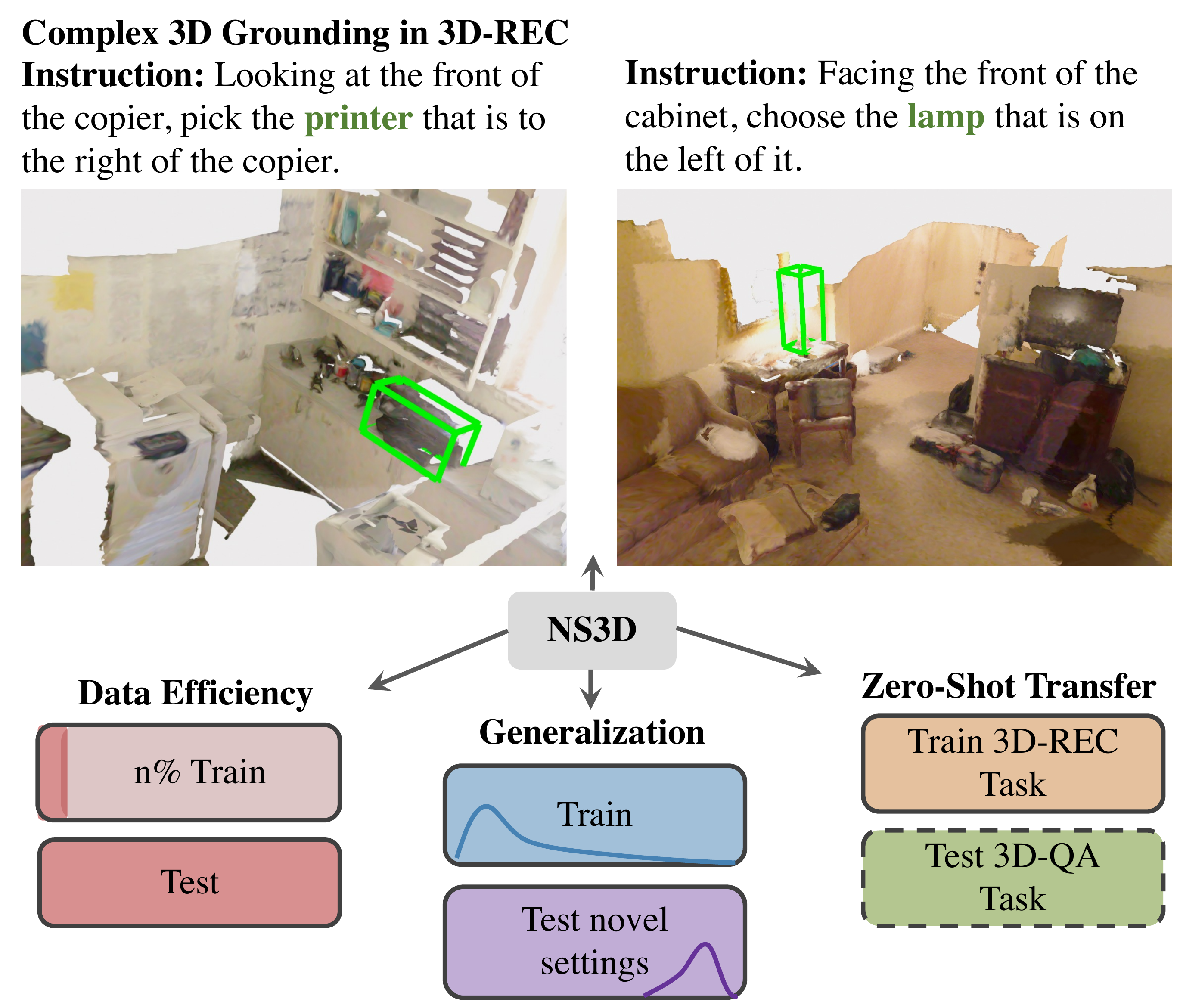}
  \caption{\model achieves grounding of 3D objects and relations in complex scenes, while showing state-of-the-art results in data efficiency, generalization, and zero-shot transfer.}
\label{fig:pull}
\vspace{-1em}
\end{figure}

Interacting with the physical world requires 3D visual understanding; it entails the ability to interpret 3D objects and relations among multiple entities, as well as reason about 3D instances in a scene from language expressions. However, due to the variability of the 3D domain, there are two prevalent challenges: the expense of annotating 3D labels and the complexity of 3D grounded language. In this paper, we tackle these two challenges on a specific task of 3D scene understanding, the referring expression comprehension (3D-REC) task. As shown in Figure~\ref{fig:pull}, in a 3D-REC task, the input contains a sentence and a 3D scene, usually given as a collection of object point clouds; the goal is to identify the correct referred object in the scene. The task is challenging: obtaining high-quality annotations for such tasks is expensive; the referring expressions often require reasoning about multiple objects, such as anchoring speaker viewpoints (\ie, facing {\it X}, select the object {\it Y} behind {\it Z}) and utilizing multiple objects in the scene as reference points.

Many prior works have studied end-to-end methods to tackle this problem \cite{yang2021sat, he2021transrefer3d, roh2022languagerefer, abdelreheem20223dreftransformer, jain2022bottom, huang2022multi, zhao20213dvg, yuan2021instancerefer, huang2021text, achlioptas2020referit3d}, jointly attending over features from language and point clouds. These methods report strong performance, but generally require large amounts of data to train and are prone to dataset biases, such as object co-occurrences. Meanwhile, the learned 3D representations cannot be directly transferred to related downstream tasks, such as 3D question answering. In addition, most prior works in 3D grounding are based on Transformers \cite{vaswani2017attention}, which reduce the set of realizable functions to a subset of reasoning tasks with binary relations \cite{vaswani2017attention, merrill2022transformers}. This has limited their ability to resolve complex 3D grounded languages, empirically leading  to a noticeable performance drop when the language contains view-dependent relations. %

To this end, we propose \model as a powerful neuro-symbolic approach to solve 3D visual reasoning tasks, with more faithful grounding of 3D objects and relations. \model first parses the referring expression from the free language form to a neuro-symbolic program form. We introduce the use of Codex~\cite{chen2021evaluating}, a large language-to-code model, for semantic parsing with a small number of prompting examples, leading to perfect identification of entities and program structures. Such program structures decompose each referring expression into a set of functional modules that are hierarchically chained together. Functional modules can perform an object-level grounding step, such as selecting the {\it bathroom vanity} from the input point clouds, and a relational grounding step, such as finding objects that are {\it behind} another reference object. This functional composition strategy can be easily extended to more complex functions that require multiple objects, such as view-dependent relation grounding. In \model, functional modules are implemented as different neural networks that take object features of the corresponding arity: \eg, object-level grounding modules take per-object features, while relation grounding modules take a set of vector encodings for each pair of objects. Importantly, \model extends prior neuro-symbolic approaches for visual reasoning \cite{mao2019neuro} by introducing modules that execute high-arity programs, such as those for relation grounding over multiple objects, especially ubiquitous in the 3D domain.

The combination of compositional structures and modular neural networks fulfills many desiderata for 3D visual reasoning (see Figure~\ref{fig:pull}). First, specializing neural modules for relations that involve multiple objects improves performance, particularly in resolving complex view-dependent referring expressions. Our approach is noticeably simpler and more effective than existing models that solve this task by fusing multiple view representations~\cite{huang2022multi}. Second, the disentangled grounding of objects and relations brings significantly improved data efficiency. Third, by following symbolic structures to compose functional modules, \model generalizes better to scenarios with unseen object co-occurrences and scene types. Fourth, the compositional nature of the functional structures and the flexibility of our Codex-based parser enables \model to zero-shot generalize to novel reasoning tasks, such as 3D visual question answering (3D-QA). Furthermore, as a byproduct of our modular approach, \model enables better interpretability, allowing attribution to where visual grounding fails and succeeds; we show in ablations that \model learns almost perfect relation grounding.

We validate \model on the ReferIt3D benchmark, which evaluates referring expression comprehension in 3D scenes, and requires fine-grained object-centric and multi-object relation grounding~\cite{achlioptas2020referit3d}. We report state-of-the-art view-dependent accuracy and comparable overall accuracy to top-performing methods. We also present results on data efficiency and generalization to unseen object co-occurrences and new scenes, with our neuro-symbolic method outperforming all prior work by a large margin. Finally, we show \model's ability to zero-shot transfer from the 3D reference task to a new 3D visual question answering task, achieving strong performance without any data in this novel setup.

To summarize, the contribution of this paper is three-fold: 1) We propose a neuro-symbolic method to ground 3D objects and relations that integrates the power of large language-to-code models and modular neural networks. 2) We introduce a neural program executor that reasons about high-arity relations as a principled solution to view-point anchoring and multi-object reference. 3) We show state-of-the-art view-dependent grounding results in 3D-REC tasks, high accuracy in data-efficient settings (a $24.5$ percent point gain from prior work with 1.5\% of data), significant improvements in generalization to different data distributions, and ability to zero-shot transfer to an unseen 3D-QA task.

\begin{figure*}[tp]
  \centering
    \includegraphics[width=1.0\textwidth]{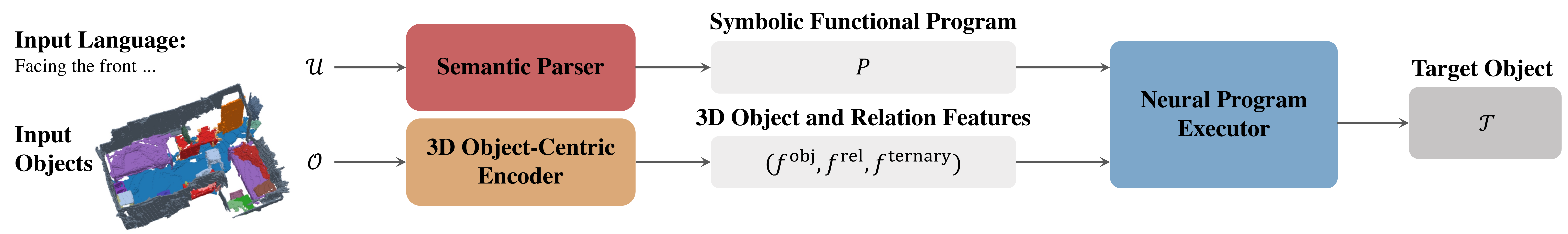}
  \vspace{-2em}
  \caption{\model is composed of three main components. a) A \textit{semantic parser} parses the input language into a symbolic program. b) A \textit{3D object-centric encoder} takes input objects and learns object, relation, and ternary relation features. c) A \textit{neural program executor} executes the symbolic program with the learned features to retrieve the target referred object.}
\label{fig:systems}
\vspace{-1em}
\end{figure*}

\section{Related Work}
\label{sec:related}
\xhdr{3D grounding.}
Many prior works that tackle the 3D-REC task employ end-to-end approaches that jointly attend over language and point clouds \cite{cai20223djcg, luo20223d, chen2022ham, chen2021d3net, chen2020scanrefer}, commonly leveraging a Transformer architecture \cite{vaswani2017attention}. These methods can be broadly categorized into two types:  object-centric ones, and ones that model the full 3D scene. Most works based on full 3D scene modeling use a detection module to create object proposals. For example, Text-guided Graph Neural Network \cite{huang2021text} conducts instance segmentation on the full scene to create candidate objects as input to a graph neural network \cite{scarselli2008graph};  InstanceRefer \cite{yuan2021instancerefer} selects instance candidates from the panoptic segmentation of point clouds; 3DVG-Transformer \cite{zhao20213dvg} uses outputs from an object proposal generation module to fully leverage contextual clues for cross-modal proposal disambiguation. The best performing work in this category, BUTD-DETR \cite{jain2022bottom}, uses box proposals from a pretrained detector and scene features from the full 3D scene to decode objects with a detection head. The Multi-View Transformer \cite{huang2022multi} separately models the scene by projecting the 3D scene to a multi-view space, to eliminate dependence on specific views and learn robust representations.

By contrast, object-centric models perform reasoning over an input set of object point clouds. ReferIt3DNet \cite{achlioptas2020referit3d} utilizes a graph convolutional network with input objects as nodes of the graph. 3DRefTransformer \cite{abdelreheem20223dreftransformer}, LanguageRefer \cite{roh2022languagerefer} TransRefer \cite{he2021transrefer3d}, and SAT \cite{yang2021sat} are Transformer-based methods that operate on language and 3D object point clouds. 3DRefTransformer \cite{abdelreheem20223dreftransformer} is an end-to-end Transformer model that incorporates an object pairwise spatial relation loss. LanguageRefer \cite{roh2022languagerefer} uses a Transformer architecture over bounding box embeddings and language embedding from DistilBert \cite{sanh2019distilbert}. TransRefer \cite{he2021transrefer3d} utilizes a Transformer-based network to extract entity-and-relation-aware representations. SAT \cite{yang2021sat} leverages 2D image semantics with a multi-modal Transformer for joint representation learning. \model lives in this category of methods, focusing on grounding objects and relations over an object-centric representation. In contrast to prior works, \model enables strong data efficiency, generalization, and zero-shot transfer to novel tasks, while not restricted to functions that Transformers can realize or constrained by the need for additional 2D data.

\xhdr{Neuro-symbolic visual reasoning methods.}
Neuro-symbolic methods have shown strong data efficiency and generalization capability in the 2D visual reasoning domain, from visual question answering to image caption retrieval \cite{nsvqa, li2020closed, chen2021meta, hudson2019learning, wu2019unified, han2019visual, mascharka2018transparency, wu2017neural, johnson2017inferring, andreas2016learning, andreas2016neural}, with the Neuro-Symbolic Concept Learner \cite{mao2019neuro} as a representative work. However, the 3D domain poses additional challenges, such as more complex, high-arity programs, required for anchoring speaker view and using multiple reference objects to resolve a referring expression. \model builds on a 3D scene-graph like representation~\cite{armeni20193d, wald2020learning}. It retains all the benefits of existing neuro-symbolic visual reasoning models, while extending them to this challenging 3D domain. %
In addition, \model sheds new lights on a broad criticism of prior neuro-symbolic methods on their use of a predefined grammar or trained parser \cite{mao2019neuro, feng2021free}: \model leverages large language-to-code models for semantic parsing \cite{chen2021evaluating}. %

\section{\model}
\label{sec:methods}

In this section, we describe \model applied to the task of referring expression comprehension (3D-REC). We set up the task following ReferIt3D~\cite{achlioptas2020referit3d}: given a set of $M$ objects in the scene $\mathcal{O} = \{O_1, ..., O_M\}$, where each object is represented as an RGB-colored point cloud of $N$ points $O_i \in \mathbb{R}^{N \times 6}$, and given an utterance $\gU$, the goal is to predict the target referred object $\mathcal{T} \in \mathcal{O}$. Due to the existence of many distractor objects in $\mathcal{O}$, it is crucial to parse the full referring expression to select the correct target.

\model is a neuro-symbolic approach that combines programmatic functional structures and modular neural networks. It consists of three main components (see Figure~\ref{fig:systems}). The first is a {\it semantic parser} that parses the input language $\gU$ into a symbolic program $P$ that resembles a hierarchical reasoning process underlying $\gU$ (Section~\ref{sec:parser}). The second is a {\it 3D feature encoder} that extracts an object-centric representation $f$ from the input point clouds of objects $\gO$ (Section~\ref{sec:encoder}). The third is a {\it neural network-based program executor} that takes the symbolic program and the learned object-centric representation, and returns the target object $\mathcal{T}$ %
(Section~\ref{sec:executor}).

\subsection{Semantic parser}
\label{sec:parser}

\begin{figure}[tp!]
  \centering\small
    \includegraphics[width=1.0\linewidth]{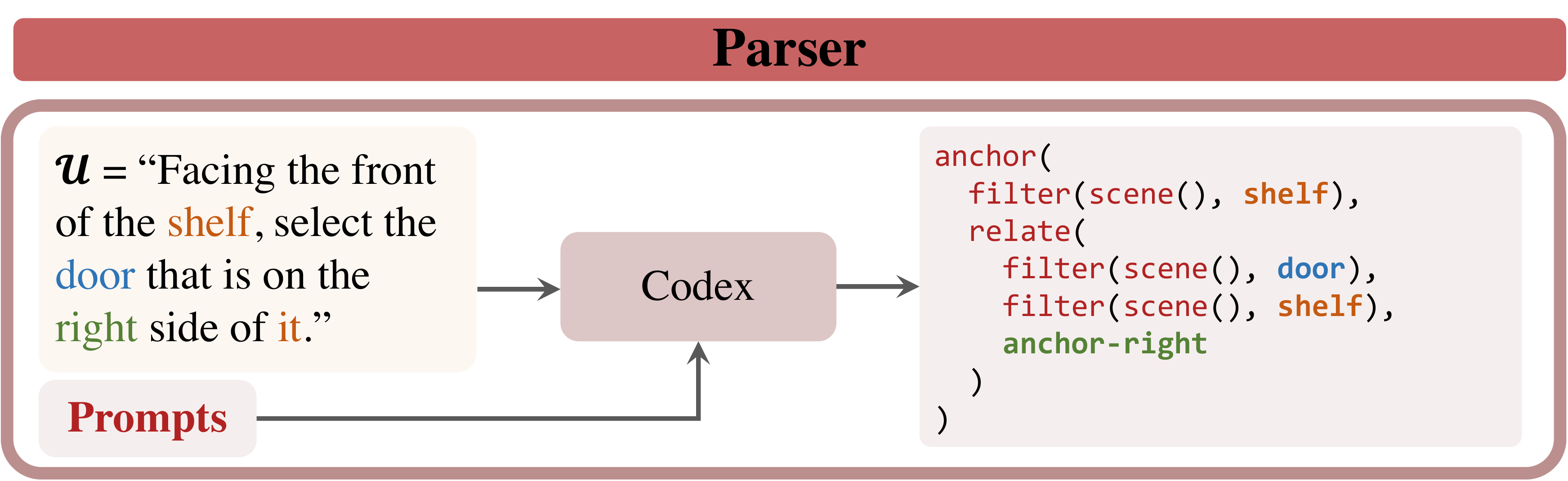}
  \caption{The \model semantic parser leverages Codex to parse input language into symbolic programs.}
\label{fig:systems_parser}
\vspace{-2.0em}
\end{figure}

The goal of the \model parser is to parse utterances $\gU$ into a symbolic program $P$ that resembles the underlying reasoning process of $\gU$. The program has a hierarchy of primitive operations defined in a minimal but powerful domain-specific language (DSL) for 3D visual reasoning tasks. Each operation is composed of a function name (\eg, \textit{anchor} or \textit{filter}), and arguments (\eg, \textit{shelf}, \textit{door}, \textit{right}). A key feature of such programs is that the output of one operation can be the input (argument) to another operation, as shown in Figure~\ref{fig:systems_parser}. Informally, the \textit{anchor} program grounds the viewpoint, the \textit{filter} program takes all the input objects in the full \textit{scene} and outputs those that are of the specified category, and the \textit{relate} program returns objects that satisfy the given relationship constraint. We include a more formal definition of these operations and the DSL in the supplementary material.

Parsing the input language into this hierarchical program allows \model to disentangle the learning of different functional modules that perform object-level or relational grounding and reasoning. These neural programs can be trained and combined in different ways.

\vspace{-1em}
\paragraph{Semantic parsing with Codex.}
In contrast to most existing neuro-symbolic reasoning frameworks, \eg,\cite{mao2019neuro}, instead of using a pretrained or jointly-trained semantic parser, we introduce the use of large language-to-code models for parsing. Specifically, we use Codex~\cite{chen2021evaluating, brown2020language} with the Synchromesh framework \cite{poesia2022synchromesh}. By specifying only a small number of examples of language input and expected programs, we gain perfect parsing capabilities across unseen categories and relations in the ReferIt3D task. Synchromesh constrains the output of Codex to be a valid, executable program, adhering to the syntactic rules of the DSL.

Using Codex as our semantic parser has two major advantages. First, it only requires a small number of example programs to achieve strong parsing accuracy, compared to defining rules for semantic parsers or training from scratch. This enables us to easily generalize to new DSLs and tasks, such as recombining the learned functional modules in a completely new way to answer visually grounded questions. Second, compared to existing works that assume a given set of visual concepts (categories and relations)~\cite{mao2019neuro,hudson2019learning}, Codex can automatically identify unseen concepts from language through its built-in knowledge, even if they never appear in the prompting examples. In our experiments, we show that Codex outperforms a T5-based parser \cite{raffel2020exploring} finetuned on the same set of prompting examples.

\subsection{3D object-centric encoder}
\label{sec:encoder}

\begin{figure}[tp!]
  \centering\small
    \includegraphics[width=1.0\linewidth]{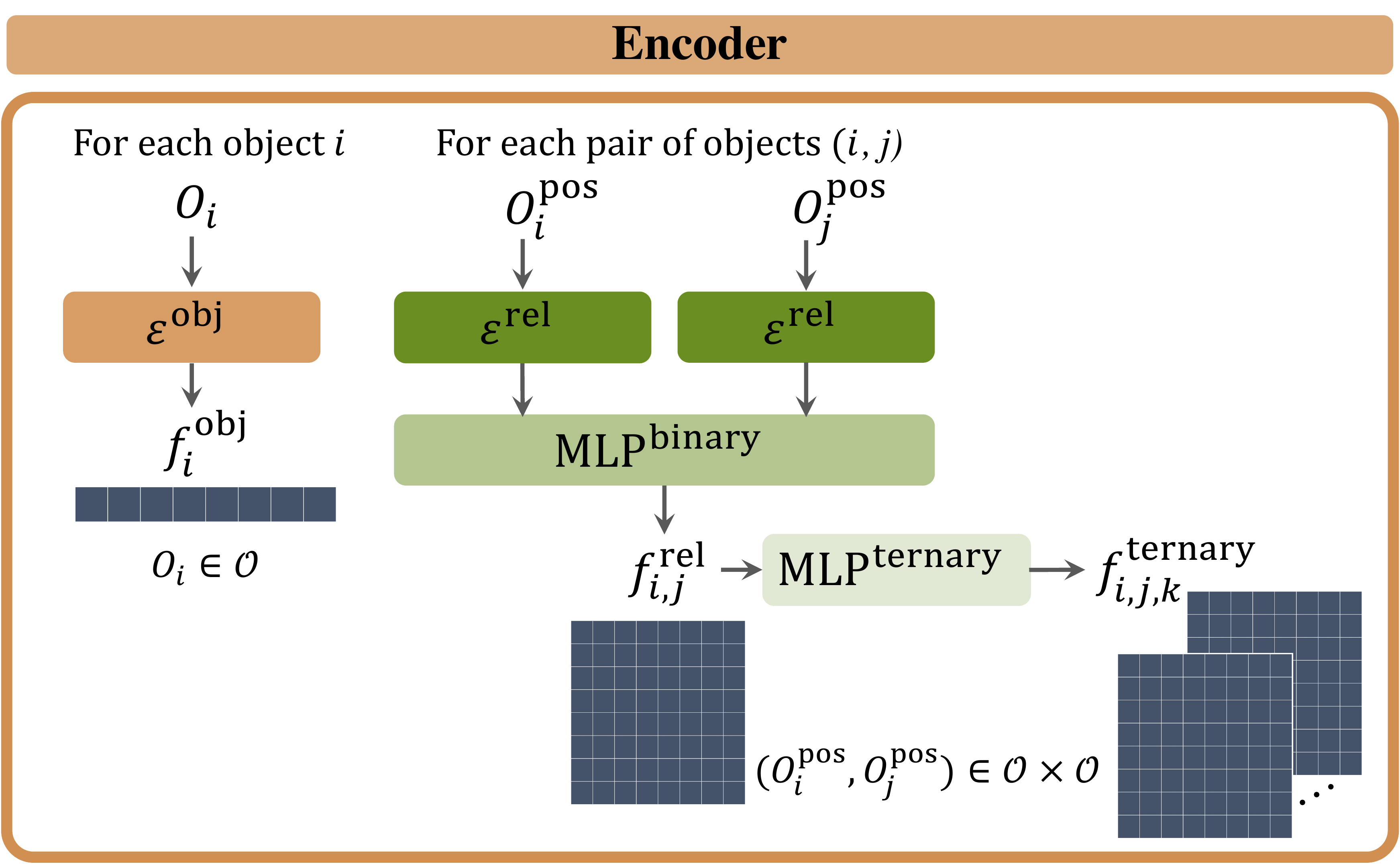}
  \caption{The \model object-centric encoder learns object, relation, and ternary relation features from input object point clouds.}
\label{fig:systems_encoder}
\vspace{-1em}
\end{figure}

\model's 3D encoder generates object-centric and relational features for each scene in a latent space for 3D grounding (see Figure~\ref{fig:systems_encoder}). Recall that the input to the encoder is a collection of object point clouds $\gO = \{O_1, O_2, \cdots, O_M\}$, where $M$ is the number of objects. For each 3D object point cloud, the encoder first extracts an object feature vector through a PointNet++ backbone $\mathcal{E}^\text{obj}$ \cite{qi2017pointnet++}. $\mathcal{E}^\text{obj}$ takes as input every object $O_i \in \mathbb{R}^{1024 \times 6}$, representing the RGB color of each point and their XYZ location in the Cartesian space, and outputs encoded features,
\vspace{-2.0em}
\begin{center}\small
\[f^\text{obj}_{i} = \mathcal{E}^\text{obj}(O_i),\ \forall O_i \in \mathcal{O} .\]
\end{center}
\vspace{-0.5em}

Next, for each pair of 3D objects $(O_i, O_j)$, \model uses a separate encoder to extract their relational feature vector. The relation encoders are designed not to share weights with the object encoders, allowing them to learn semantically relevant features for relational reasoning. Specifically, \model first encodes each object with a different PointNet++ network $\gE^\text{rel}$. $\gE^\text{rel}$ takes in only the XYZ positions of object point clouds $O^\text{pos}_i \in \mathbb{R}^{1024 \times 3}$, as we are interested in modeling the spatial relations between objects rather than object types. Next, \model concatenates the per-object encoding for $O_i$ and $O_j$ and applies a small 2-layer multi-layer perceptron $\mathrm{MLP}^\text{binary}$ to extract relational features for the object pair:
\vspace{-2.0em}
\begin{center}\small
\[f^\text{rel}_{i,j} = \mathrm{MLP}^\text{binary}\left( \mathrm{concat}\left( \gE^\text{rel}(O^\text{pos}_i), \gE^\text{rel}(O^\text{pos}_j) \right) \right), \]
\end{center}
\vspace{-0.5em}
where $\mathrm{concat}$ denotes the concatenation operation for two vectors. In our design, $\gE^\text{rel}$ is a shallower network that uses a sparser amount of samples than the object feature encoder $\gE^\text{obj}$, which requires more fine-grained encoding of point clouds to classify categories. %

We also model ternary relations as seen in ReferIt3D (\eg, a vector embedding for each triple of objects $(O_i, O_j, O_k)$). Specifically, we propose using the encoder to also extract a ternary feature $f^\text{ternary}_{i,j,k}$. As both the binary-relation features and ternary-relation features focus on encoding spatial relationships among objects, \model shares the underlying PointNet encoder for them. This reduces the time and memory cost for high-arity inference. Mathematically,
\vspace{-2.0em}
\begin{center}\small
\begin{align*}
g^\text{ternary}_{i,j} = &  \mathrm{MLP}^\text{ternary}\left( f^\text{rel}_{i, j} \right), \\
f^\text{ternary}_{i,j,k} = & \mathrm{concat}\left(
    g^\text{ternary}_{i,j}, g^\text{ternary}_{j,k}, g^\text{ternary}_{i,k}
\right),
\end{align*}
\end{center}
\vspace{-0.5em}
where $\mathrm{MLP}^\text{ternary}$ is another 2-layer multi-layer perceptron that shares the same architecture as $\mathrm{MLP}^\text{binary}$.

\subsection{Neural program executor}
\label{sec:executor}

\begin{figure}[tp!]
  \centering\small
    \includegraphics[width=1.0\linewidth]{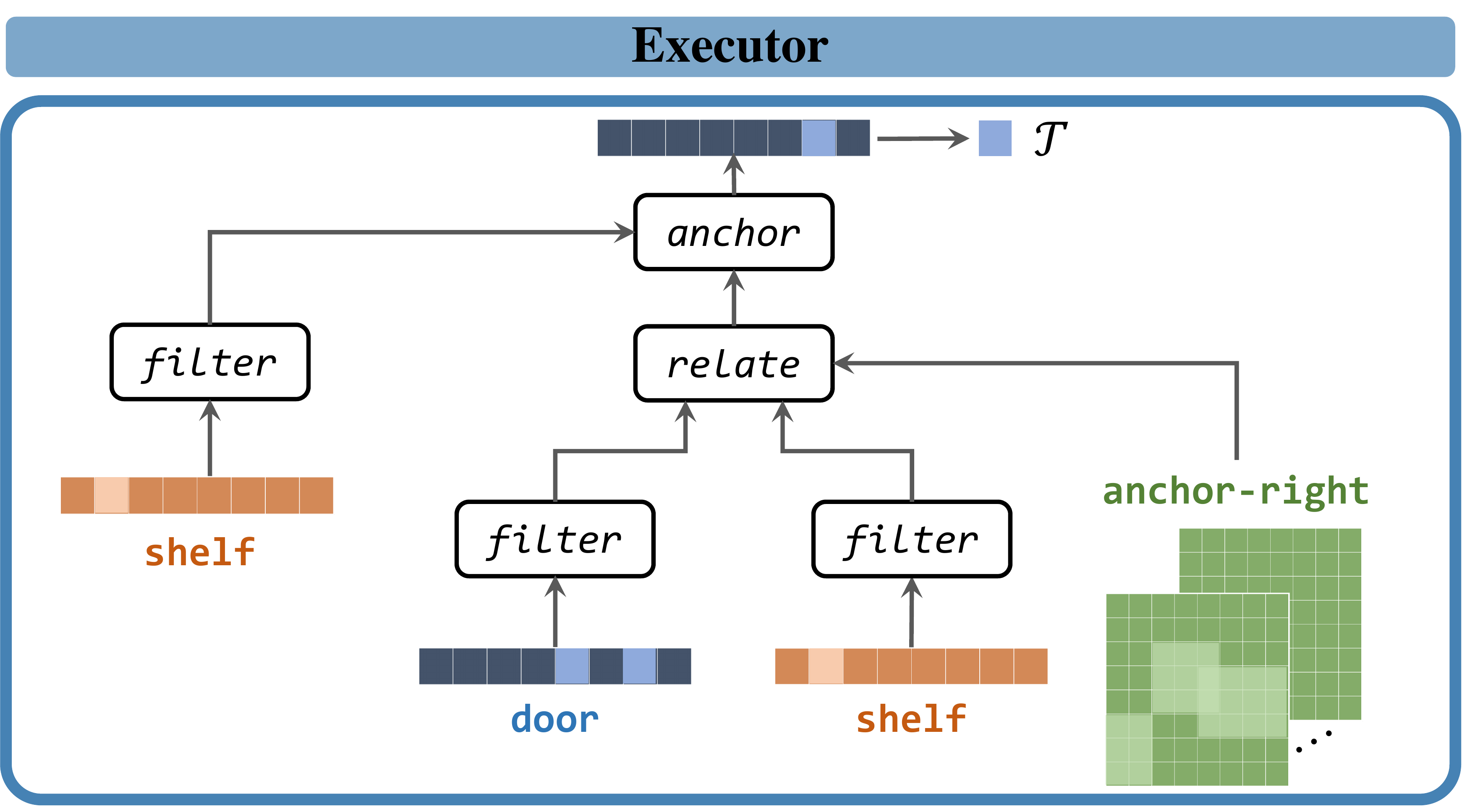}
  \caption{The \model neural program executor executes the symbolic program recursively with the learned 3D features, and returns the target referred object $\mathcal{T}$.}
\label{fig:systems_executor}
\vspace{-1em}
\end{figure}

The neural program executor takes the parsed program $P$ and the learned representation $(f^\text{obj}, f^\text{rel}, f^\text{ternary})$ for each scene as input, and executes the program based on the 3D representation, returning the target object being referred to. At a high-level, \model follows the hierarchical structure in $P$ and executes the program recursively (see Figure~\ref{fig:systems_executor}). Here, we describe the neural implementation for each operation.

The key representation we will be using during program execution is the object score vector, which is a vector of length $M$ (the number of objects in the scene), indicating whether an object has been selected or not. For example, semantically, the operation {\it filter} takes an input set of objects being selected and an object category, and outputs a subset of input objects belonging to the specified category. Both the input and output of the {\it filter} operation will be represented as such object score vectors. For numerical stability, such scores are represented in the log space. One interpretation is that each entry $v_i$ in a score vector represents the log probability of object $O_i$ being selected. %

\vspace{4pt}
\noindent {\it \textbf{scene}() $\rightarrow$ $y$}: the scene operation returns a object score vector representing ``all objects in the scene.'' Recall that the values are in the log space; therefore, $y_i = 0$, for all $i \in \{1,2,\cdots,M\}$.

\vspace{4pt}
\noindent {\it \textbf{filter}($x$, $c$) $\rightarrow$ $y$}: the filter operation takes an input object score vector $x$, an object category $c$, and returns a new object score vector, selecting objects that are in $x$ and belongs to category $c$. Therefore, we first compute $\textit{prob}^\text{c}_i = \mathrm{MLP}^\text{c}\left( f^\text{obj}_i \right)$, which is a score for object $i$ belonging to category $c$, where $\mathrm{MLP}^\text{c}$ is a mapping (specialized for $c$) from the dimension of $f^\text{obj}_i$ to dimension 1. Next, we merge it with the input object score vector $x$. Overall,
\vspace{-2.0em}
\begin{center}
\small
\[ y_i = \min\left( x_i, \textit{prob}^\text{c}_i \right) = \min\left( x_i, \mathrm{MLP}^\text{c}\left( f^\text{obj}_i \right) \right). \]
\end{center}
\vspace{-0.5em}

\vspace{4pt}
\noindent {\it \textbf{relate}($x^t$, $x^r$, $\textit{rel}$) $\rightarrow$ $y$}:
the \textit{relate} program takes as input two sets of objects, target objects $x^t$ and reference objects $x^r$, as well as a relational concept $\textit{rel}$, and outputs target objects that satisfy the specified relation. As an example, the expression ``the {chair beside} the shelf'' will be parsed into the program {\it relate(filter({chair}), filter(shelf), {beside})}\footnote{For conciseness, we have used {\it filter(shelf)} as a short-hand notation for {\it filter(scene(), shelf)}.}. In this case, $x^t$ will be the {\it filter} result for {\it {chair}}, while $x^r$ will be the {\it filter} result for {\it shelf}. The {\it relate} operation classifies whether each pair of objects satisfy the relation $\textit{rel}$, and selects the objects in $x^t$ that have relation $\textit{rel}$ with objects in $x^r$:
\vspace{-2.0em}
\begin{center}
\small
\begin{align*}
    \textit{prob}^\text{rel}_{i, j} = & \mathrm{MLP}^{\text{rel}}(f^\text{rel}_{i, j})\\
    y_i = & \min\left( x^t_i, \sum_j \textit{sx}(x^r)_j \cdot \textit{prob}^\text{rel}_{i, j} \right),
\end{align*}
\end{center}
\vspace{-0.5em}
where $\mathrm{MLP}^{\text{rel}}$ is a linear layer with scalar output and specialized for concept $\textit{rel}$, and $\textit{sx}$ is the softmax function applied to $x_r$. The $\sum_j$ operator can be interpreted as a ``soft'' selection of the $j^*$-th row in the relation matrix $\textit{prob}^\text{rel}$, where $j^* = \arg\max x^r$, the index of the referred object.

\vspace{4pt}
\noindent {\it \textbf{ternary\_relate}($x^t$, $x^{r1}$, $x^{r2}$, $\textit{trel}$) $\rightarrow$ $y$}: we propose to extend the formulations above to handle object relationships that involve more than two objects. In this case, $x^{r1}$ and $x^{r2}$ are two reference objects. In ReferIt3D~\cite{achlioptas2020referit3d}, there are two types of ternary relationships: spatial ternary relations (\eg, {\it between}), and view-dependent relations. Both are resolved with this operation. As an example, the sentence ``Facing the front of the shelf, select the door that is on the right side of it.'' yields the program {\it anchor(filter(shelf), ...)}. Internally, such {\it anchor} operation will be handled as a ternary relation function: {\it ternary\_relate(
  filter(door), filter(shelf), filter(shelf), anchor-right
)}. The two reference objects (the reference for the relation ``right'' and the anchor for ``facing'') are the same. Notably, \model's ternary operation can be generalized as a principled solution to any high-arity relations that can be executed based on learned features of the corresponding arity. In our ternary case, it is executed as the following:
\vspace{-1.5em}
\begin{center}\small
\begin{align*}
& \textit{prob}^\text{trel}_{i, j, k} = \mathrm{MLP}^{\text{trel}}(f^\text{ternary}_{i, j, k})\\
& y_i = \min\left( x^t_i, \sum_j \sum_k \textit{sx}(x^{r1})_j \cdot \textit{sx}(x^{r2})_k \cdot \textit{prob}^\text{trel}_{i, j, k} \right).
\end{align*}
\end{center}
\vspace{-0.5em}

The \model neural program executor composes the above operations and outputs the final object score vector, whose maximum-valued index represents the referred object $\mathcal{T}$. %

\subsection{Training}
Modules in \model can be trained end-to-end with only the groundtruth referred objects as supervision; each can also be trained individually whenever additional labels are available. In this paper, we use a hybrid training objective similar to prior works~\cite{achlioptas2020referit3d,jain2022bottom}. Specifically, we use the groundtruth object category to compute a per-object classification loss $\mathcal{L}_\textit{oce}$ (applied to all $\textit{prob}^\text{c}$, where $c$ is the category) and the groundtruth final target object to compute a per-expression loss $\mathcal{L}_\textit{tce}$. Both loss functions are standard cross-entropy losses. The final total loss, with $\alpha = 1$ in our experiments, is:
$\mathcal{L}_\textit{total} = \mathcal{L}_\textit{oce} + \alpha (\mathcal{L}_\textit{tce}).$ Practically, we perform a two-stage training: we first pretrain the model with $\mathcal{L}_{oce}$ until convergence, and then train with the full loss. %

As a byproduct of \model's modular structure, we gain improved model interpretability. We can validate whether specific programs yield correct outputs at each stage. This can help determine what types of additional data will be valuable. In our experiments, we find that object categorization is the most challenging part of the 3D-REC task.

\model does not need the full scene point cloud as its input, and instead only explicitly models a given object set $\mathcal{O}$. Therefore, we can train on scenes with a small number of objects ($10$ objects in our experiments) and directly test on scenes with much more objects ($88$ objects maximum in the test set). This improves training efficiency and reduces the need for annotated 3D objects, which are expensive to acquire in 3D domains. It also enables generalization to more cluttered and complex scenes. In all of our experiments, we train \model on scenes with a sparse amount of objects and see no performance drop compared to training with a dense train set. By contrast, baseline methods yield significantly decreased performance under this setting; we present these results in the supplementary material.
\section{Experiments}
\label{sec:experiments}
We evaluate \model and compare it to prior work on the ReferIt3D benchmark \cite{achlioptas2020referit3d}, a 3D referring expression comprehension (3D-REC) task. We specifically focus on the SR3D setting, which tackles spatially-oriented object referential language in 3D scenes. In Section~\ref{sec:referential_task}, we compare performance against baselines and report ablations. In Section~\ref{sec:data_efficiency} and Section~\ref{sec:generalization}, we show experiments on data efficiency and generalization. Finally, in Section~\ref{sec:transfer}, we present \model's ability to zero-shot transfer to a 3D-QA task.

\subsection{3D referring expression comprehension}
\label{sec:referential_task}
\begin{figure}[tp!]
  \centering\small
    \includegraphics[width=1\linewidth]{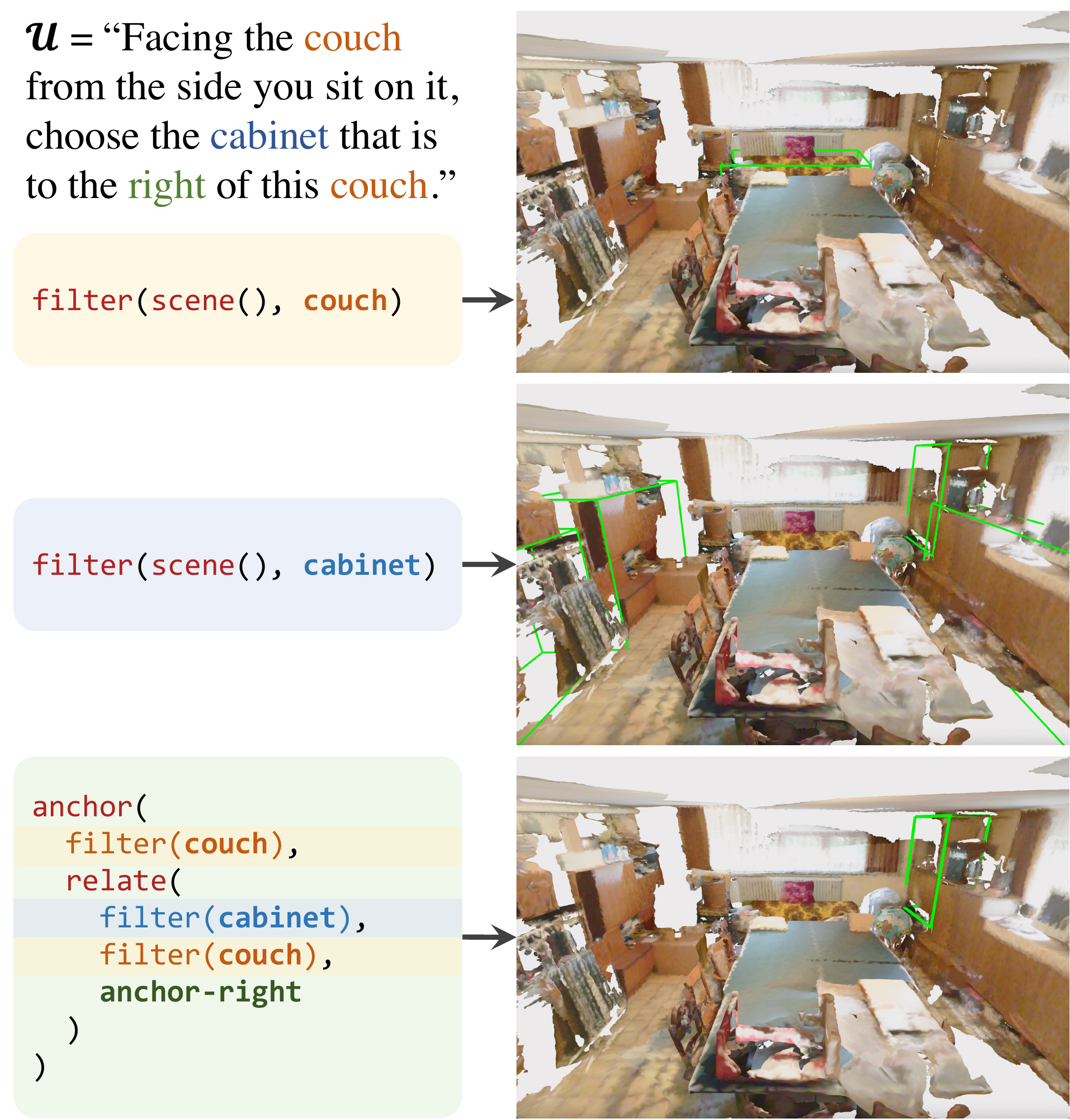}
  \caption{Example of \model's execution trace on a view-dependent 3D-REC task. \model returns the correct target cabinet object.} %
\label{fig:vd}
\vspace{-1.5em}
\end{figure}

In the ReferIt3D task, the input is a set of point clouds, one for each object in the scene, as well as the utterance, and the target object is assumed to be unique. Therefore, models can be evaluated by the accuracy of selecting the correct object.
Figure~\ref{fig:vd} shows an example task instance in ReferIt3D and its \model execution trace.

\vspace{4pt}
\xhdr{Results.}
In Table~\ref{table:main_comparison}, we compare \model to baselines. We group methods into two categories: methods that only use the per-object point clouds and methods that model the full scene. The results show that we outperform other object-centric methods and achieve comparable performance with top-performing methods. In addition, we demonstrate state-of-the-art view-dependent accuracy compared to all prior work, with an improvement of 3.6\% against the top-performing baseline. \model shows close performance between overall and view-dependent accuracy, while all other methods yield a large gap. Note that unlike models such as MVT~\cite{huang2022multi} that explicitly transform and encode point clouds from multiple views to improve multi-view performances, our model simply uses a general high-arity neural network.

\vspace{4pt}
\xhdr{Ablations.}
In Table~\ref{table:ablation}, we first discuss the performance of relational grounding modules. Specifically, since the ReferIt3D dataset does not contain groundtruth labels for object relations, we study the performance of relation grounding modules by evaluting \model performance using the groundtruth object classification for all \textit{filter} operations. We see that \model achieves almost perfect performance on the task, indicating that it learns relations and high-arity relations well. Our neuro-symbolic approach allows for such diagnostics of model performance: the primary challenge to \model is object classification, which can potentially be improved with more object labels.

\begin{table}[tp]
\centering\small
\begin{sc}
\begin{tabular}{lll}
    \toprule
     & Overall & View-dep. \\
    \midrule
    \multicolumn{3}{l}{\textbf{- Without 3D scene modeling}} \\
    \midrule
    \model (ours) & $\mathbf{0.627}$ & $\mathbf{0.620}$ \\
    SAT \cite{yang2021sat} & $0.579$ & $0.492$ \\
    TransRefer \cite{he2021transrefer3d} &	$0.574$  &	$0.499$  \\
    LanguageRefer \cite{roh2022languagerefer} &	$0.560$ &	$0.492$ \\
    3DRefTransformer \cite{abdelreheem20223dreftransformer} &	$0.470$  &	$0.443$ \\
    ReferIt3D \cite{achlioptas2020referit3d} &	$0.408$ &	$0.392$   \\
    \midrule
    \multicolumn{3}{l}{\textbf{+ With 3D scene modeling}} \\ %
    \midrule
    BUTD-DETR \cite{jain2022bottom} & $\mathbf{0.670}$\footnotemark{}   & $0.530$  \\
    MVT \cite{huang2022multi} & $0.645$ & $\mathbf{0.584}$  \\
    3DVG-Transformer \cite{zhao20213dvg} &	$0.514$  &	$0.446$ \\
    InstanceRefer \cite{yuan2021instancerefer} &	$0.480$ &	$0.454$ \\
    Text-Guided-GNNs \cite{huang2021text} &	$0.450$ &	$0.458$ \\
    \bottomrule
  \end{tabular}
  \vspace{-0.5em}
  \caption{\model yields the highest overall accuracy on the SR3D task among object-centric methods, and state-of-the-art view-dependent accuracy across all methods.}
  \label{table:main_comparison}
\end{sc}
\vspace{-0.5em}
\end{table}

\begin{table*}[tp!]
\centering\small
\setlength{\tabcolsep}{5pt}
\begin{sc}
\begin{tabular}{lllllllllllllll}
    \toprule
    & \multicolumn{2}{c}{$0.5\%$} & & \multicolumn{2}{c}{$1.5\%$}& & \multicolumn{2}{c}{$2.5\%$} & & \multicolumn{2}{c}{$5\%$} & & \multicolumn{2}{c}{$10\%$} \\
    \cmidrule{2-3} \cmidrule{5-6} \cmidrule{8-9} \cmidrule{11-12} \cmidrule{14-15}
    & All & V-dep. & & All & V-dep. & & All & V-dep. & & All & V-dep. & & All & V-dep.\\
    \midrule
    \model + full (Ours) & $\mathbf{0.503}$ & $\mathbf{0.395}$ & & $\mathbf{0.576}$ & $\mathbf{0.468}$ & & $\mathbf{0.587}$ & $\mathbf{0.505}$ & & $\mathbf{0.597}$ & $\mathbf{0.505}$ & & $\mathbf{0.612}$ & $\mathbf{0.552}$ \\
    \model (Ours) & $0.426$ & $0.375$ & & $0.520$ & $0.424$ & & $0.556$ & $0.483$ & & $0.591$ & $0.493$ & & $0.600$ & $0.527$ \\
    BUTD-DETR \cite{jain2022bottom} &  $ 0.083$ & $0.089$ & & $0.158$ & $0.138$ & & $0.259$ & $0.223$ & & $0.395$ & $0.302$ & & $0.528$ & $0.420$ \\
    MVT \cite{huang2022multi} & $0.161$ & $0.118$ & & $0.275$ & $0.199$ & & $0.322$ & $0.270$ & & $0.380$ & $0.375$ & & $0.491$ & $0.426$ \\
    SAT \cite{yang2021sat} & $0.172$ & $0.149$ & & $0.260$ & $0.254$ & & $0.298$ & $0.273$ & & $0.330$ & $0.309$ & & $0.362$ & $0.334$ \\
    TransRefer \cite{he2021transrefer3d} & $0.188$ & $0.152$ & & $0.268$ & $0.233$ & & $0.305$ & $0.278$ & & $0.362$ & $0.380$ & & $0.390$ & $0.378$ \\
    \bottomrule
  \end{tabular}
  \vspace{-0.5em}
  \caption{Data efficiency results of \model compared to prior works, with 0.5\%, 1.5\%, 2.5\%, 5\%, and 10\% of train data. We report two variations of \model, with object classification pretrained on the full dataset and pretrained on the specified data-efficient train set.}
  \label{table:efficiency}
\end{sc}
\vspace{-1.5em}
\end{table*}

We next explore the importance of separating object encoders $\mathcal{E}^\text{obj}$ and $\mathcal{E}^\text{rel}$. Having separate object and relation features allows each to specialize in different goals: object classification and relational reasoning. We see that overall performance decreases by $4.6\%$ if they share the same feature encoder. We additionally present results on using an incorrect number of arguments for executing high-arity queries (\ie, spatial ternary relations and view-dependent relations) by removing the second reference object. It leads to a $5.8\%$ drop in view-dependent accuracy, which supports the importance of high-arity modules.

\begin{table}[tp!]
\centering\small
\begin{sc}
\begin{tabular}{llll}
    \toprule
     & Overall  & View-dep. \\
    \midrule
    \model w/ gt obj. cls. & $0.969$ & $0.823$ \\
    \model w/o sep. feat. &  $0.581$ & $0.512$ \\
    \model w/o ternary arg. &  $0.609$ & $0.562$ \\
    \midrule
    \model (Full) & $0.627$ & $0.620$ \\
    \bottomrule
  \end{tabular}
  \caption{Ablation on \model with groundtruth object classification results in \textit{filter}, without separation of object and relation features, and without leveraging the correct arity for ternary operations.}
  \label{table:ablation}
\end{sc}
\vspace{-1.5em}
\end{table}

\footnotetext{We note that BUTD-DETR operates on a more constrained setting, assuming 485 classes instead of the full 607 classes in ReferIt3D.}

\subsection{Data efficiency}
\label{sec:data_efficiency}

We report experiments on data efficiency compared to four top-performing prior work on ReferIt3D, two object-centric methods (SAT \cite{yang2021sat} and TransRefer \cite{he2021transrefer3d}) and two methods that model the full 3D scene (BUTD-DETR \cite{jain2022bottom} and MVT \cite{huang2022multi}). We test on 0.5\% (329 examples), 1.5\% (987 examples), 2.5\% (1,646 examples), 5\% (3,292 examples), and 10\% of data (6,584 examples) in the train set, with the same full test set. We note that BUTD-DETR \cite{jain2022bottom} uses pretrained object classification results on the full ScanNet dataset \cite{dai2017scannet}, while others do not. Hence in Table~\ref{table:efficiency}, we report \model's performance on both settings: pretrained object classification on the full ReferIt3D train set (\model + Full), and on the smaller train set only (\model). %

\begin{figure}[tp!]
  \centering
  \includegraphics[width=1.0\linewidth]{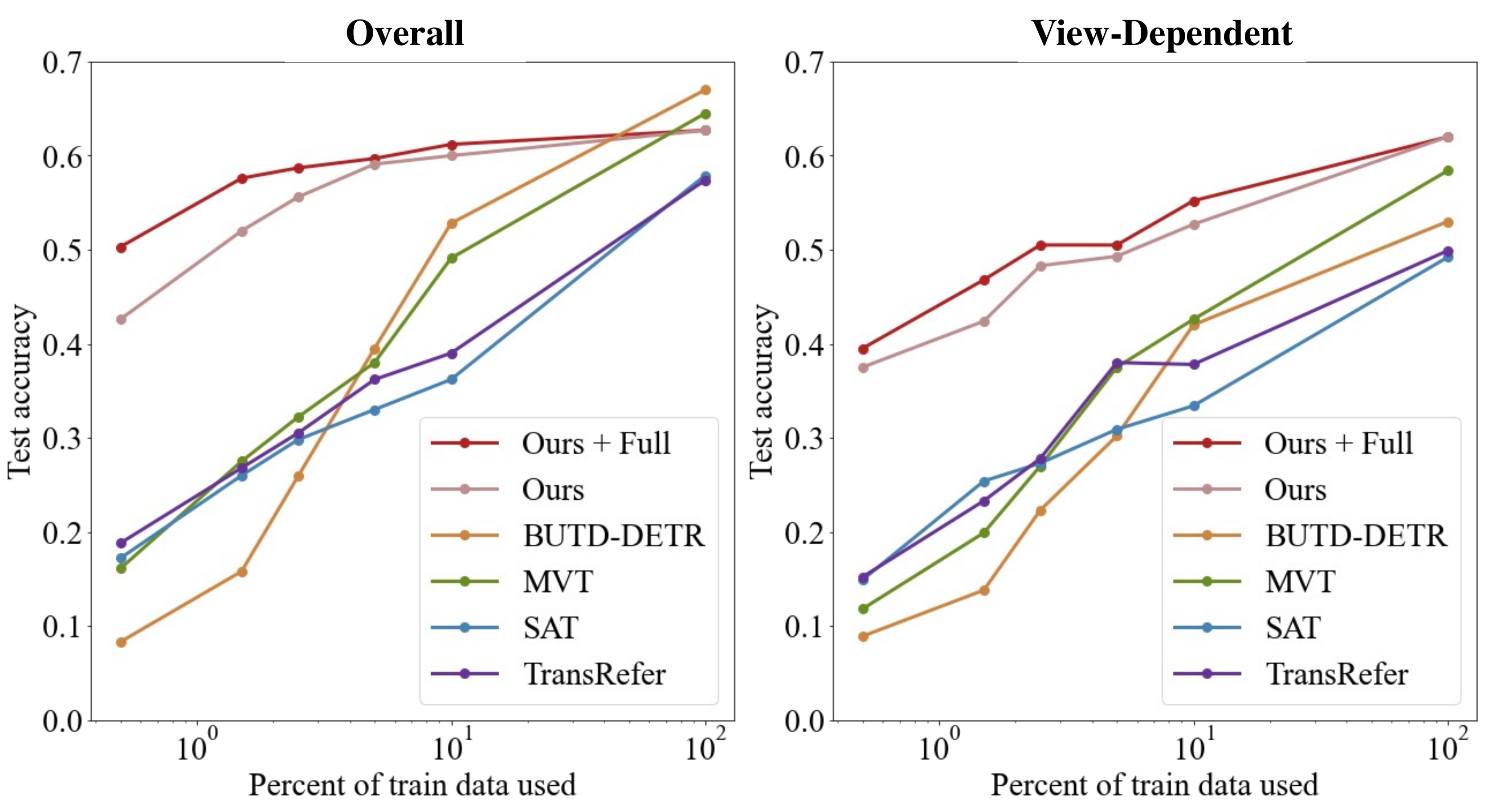}
  \caption{\model outperforms prior works by a large margin with 0.5\%, 1.5\%, 2.5\%, 5\%, and 10\% of train data.}
\label{fig:data_eff}
\vspace{-1em}
\end{figure}

Shown in Figure~\ref{fig:data_eff}, we see that in both settings, our neuro-symbolic approach significantly outperforms prior work across all data-efficient settings, achieving only small drops in accuracy compared to training on the full train set. \model yields 52.0\% accuracy with just 1.5\% of the train data, with 987 examples only, while all other baselines report accuracy lower than 27.5\%. We see this trend persist across data-efficient settings. \model sees only a 3.6\% gap when using 5\% vs 100\% of data, while all other methods decrease in performance significantly. \model's accuracy of $59.1\%$ at 5\% train data is higher than that of baselines at 100\% train data, aside from BUTD-DETR \cite{jain2022bottom} and MVT \cite{huang2022multi}. This is a significant improvement in the 3D domain, where data annotation is especially labor intensive and expensive.

\subsection{Generalization}
\label{sec:generalization}
We present two additional generalization settings and compare our model against prior work in Table~\ref{table:generalization}. The first setting (PAIRS) evaluates performance on unseen object-relation-object pairs. The referring expressions in the train set include the top 5 percent of object-relation-object pairs: \ie, the referred object category, relation type, and the reference object category (\eg, chair-closest-door). The test set contains the bottom 95 percent of object-relation-object pairs in the long-tailed distribution. \model does not see a noticeable performance drop, while methods that encode dataset bias by attending over all objects regardless of the functional structures perform poorly in evaluation.

The second setting (SCENE) evaluates model performance on an unseen scene type. The train set includes train examples with all scene types aside from that of ``living room'', while the test set only contains examples in living rooms. This is an important generalization setting, as we often want to evaluate models on new environments, without having to additionally label examples with every new scene type. \model outperforms all prior work in this setting\footnote{We note that view-dependent accuracy for scene generalization has high variance due to the small amount (36 examples) of test view-dependent examples in living room scenes.}.

\begin{table}[tp!]
\small\centering\small
\setlength{\tabcolsep}{4pt}
\begin{sc}
\begin{tabular}{llllll}
    \toprule
    & \multicolumn{2}{c}{Pairs}  & & \multicolumn{2}{c}{Scene} \\
    \cmidrule{2-3} \cmidrule{5-6}
     & All & V-dep. & & All & V-dep \\
    \midrule
    \model + full (Ours) & $\mathbf{0.612}$ & $\mathbf{0.635}$ & & $\mathbf{0.563}$ & $0.583$ \\
    \model (Ours) & $0.599$ &  $0.620$ & & $0.544$ & $\mathbf{0.611}$ \\
    BUTD-DETR \cite{jain2022bottom} & $0.440$ & $0.423$ & & $0.515$ & $0.583$  \\
    MVT \cite{huang2022multi} & $0.420$ & $0.353$ & & $0.502$ & $0.500$ \\
    SAT \cite{yang2021sat}  & $0.359$ & $0.380$ & & $0.451$ & $0.500$ \\
    TransRefer \cite{he2021transrefer3d}  & $0.322$ & $0.344$ & & $0.384$ &  $0.361$ \\
    \bottomrule
  \end{tabular}
  \vspace{-0.5em}
  \caption{Generalization performance to unseen object co-occurrence pairs and scene. \model outperforms all baselines.}
  \label{table:generalization}
\end{sc}
\vspace{-1.5em}
\end{table}

\subsection{Zero-shot transfer}
\label{sec:transfer}
\begin{figure}[tp!]
  \centering\small
    \includegraphics[width=1.0\linewidth]{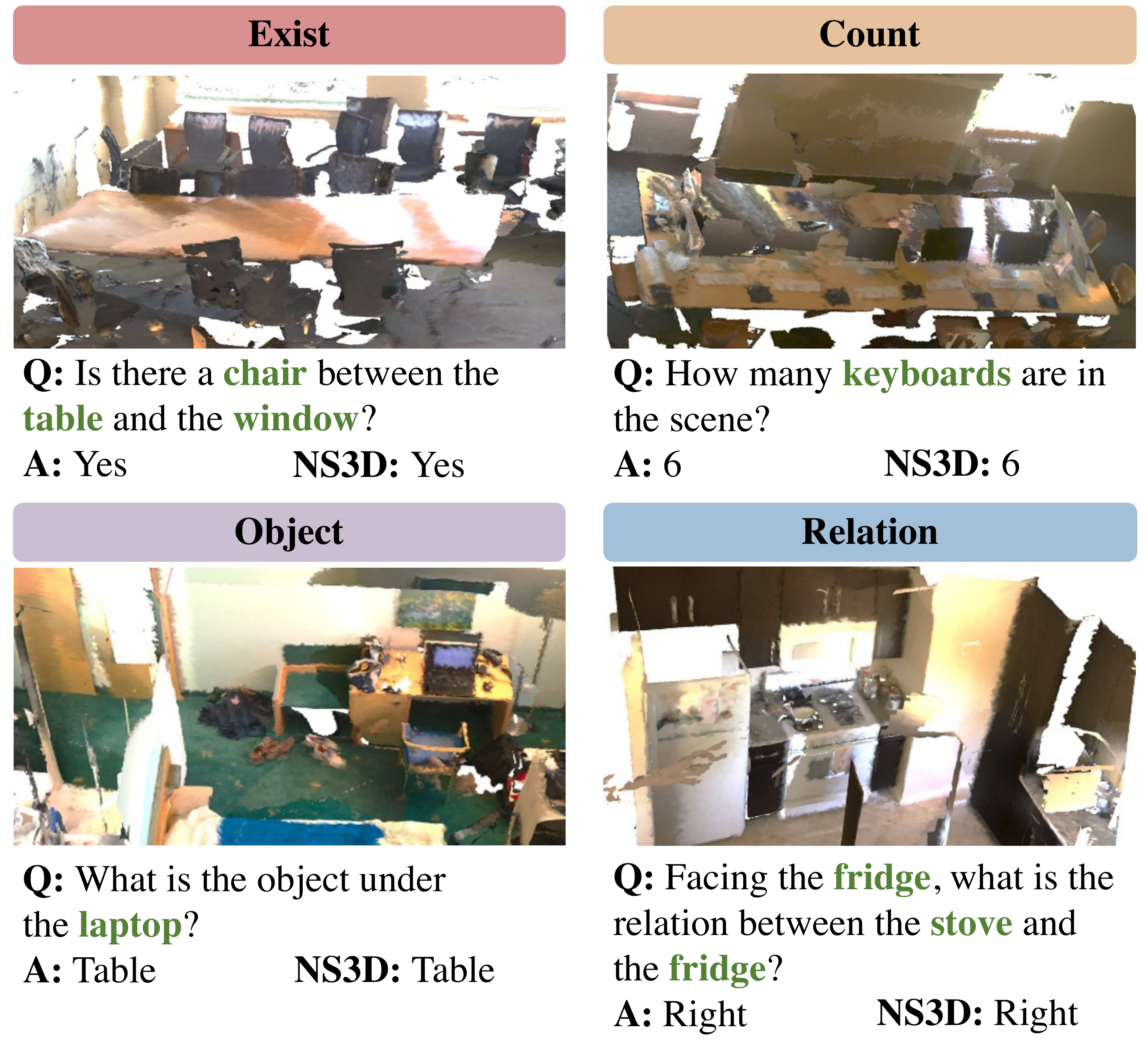}
    \vspace{-2.5em}
  \caption{Examples of 4 questions types from the 3D-QA task.}
\label{fig:qa}
\vspace{-2.0em}
\end{figure}

Finally, we showcase \model's ability to zero-shot transfer to a new 3D question answering task (3D-QA). Since there are no existing closed-vocabulary 3D-QA datasets in the same domain, we have manually created a small evaluation set of $50$ examples for 3D-QA, where the input is a set of objects in the scene, $\mathcal{O} = \{O_1, ..., O_M\}$, and a question $\gQ$. In contrast to the 3D-REC task, where the output is the target object, the output for 3D-QA is an answer in text form (the vocabulary contains all categories, relations, Yes/No, and integers). The dataset consists of four main types of questions; see Figure~\ref{fig:qa} for examples of each type. The first are exist-typed questions, which ask whether an object of the specified class and relation exists. The second are count-typed questions, which ask for the number of objects that satisfies the specification. The third are object-typed questions, which ask for object categories, and the last are relation-typed questions, which ask for the relationship between the specified objects. Each type of question has view-dependent and ternary relation variants.

In the semantic parsing stage, \model parses the new input questions into symbolic programs by specifying only a handful of prompts for our Codex-based parser (10 sentence-program pairs). By simply specifying one prompt for each type of expected program structure, we gain perfect parsing capabilities of this new program structure. In the 3D feature encoding stage, \model can directly re-use learned object and relation grounding modules from 3D-REC for 3D-QA. The 3D object-centric features are the same across both tasks.

The new functional modules introduced in \model for 3D-QA task %
output text answers. The \textit{query\_exist} operation is implemented as max over a threshold, and the \textit{query\_count} operation as the sum over a threshold, both based on the object score vector. The \textit{query\_object} and \textit{query\_relation} operations return the category or relation label with the highest prediction probability across labels. Formal definitions for each operation are described in the supplementary material. Note that all modules require no additional training, and are built through composing learned models from 3D-REC.

We show that \model can zero-shot transfer across tasks in 3D, with \textit{no additional finetuning or training} of neural networks required. In Table~\ref{table:qa}, we report accuracy, calculated as the exact match of the text output, for overall performance and for each question type. We compare against \model with a finetuned T5 conditional generation model \cite{raffel2020exploring} as its semantic parser (\model + T5), instead of with Codex (\model + Codex). The T5 model is pretrained then finetuned on the same set of examples that Codex received. \model with Codex outperforms \model with the finetuned T5 model by a large margin, due to T5's inability to generalize to new words outside of its small train set as well as errors in parsing text into syntactically and semantically correct programs. We also report accuracy from a randomly initialized \model model, showing that executing programs with learned modules is indeed significantly more successful in this task. We show more qualitative examples in the supplementary material.

\begin{table}[tp!]
\centering\small
\begin{sc}
\begin{tabular}{llllll}
    \toprule
     & All & Exist & Count & Obj & Rel \\
    \midrule
    \model + Codex & $\mathbf{0.68}$ & $\mathbf{0.80}$ & $\mathbf{0.67}$ & $\mathbf{0.60}$ & $\mathbf{0.60}$ \\
    \model + T5 & $0.30$ & $0.40$ & $0.13$ & $0.40$ & $0.30$ \\
    Random & $0.16$ & $0.40$ & $0.07$ & $0.00$ & $0.10$ \\
    \bottomrule
  \end{tabular}
  \vspace{-0.5em}
  \caption{\model's zero-shot transfer performance on the 3D-QA task, with comparison of semantic parsers (Codex vs T5), and with a randomly initialized model as baseline.}
  \label{table:qa}
\end{sc}
\vspace{-1.0em}
\end{table}

\vspace{-0.5em}
\section{Conclusion}
\label{sec:conclusions}
\vspace{-0.3em}
We have presented \model, a neuro-symbolic model for 3D grounding that leverages compositional programs and modular neural networks to solve complex 3D-REC tasks. It enables strong data-efficiency, generalization to novel data distributions, and zero-shot transfer of 3D knowledge to a new 3D-QA task. We show that we can integrate large language-to-code models with modular neural networks, and accurately parse language into symbolic programs for visual reasoning. We also present a neural program executor that implements high-arity operations effectively to ground complex semantic forms, such as view-dependent anchoring. Together, these components of \model form a powerful model for 3D visual understanding. As a future direction, combining \model with strong object localization models can potentially enable learning directly from 3D scenes.

\vspace{-1.0em}
\paragraph{Acknowledgments.}
This work is in part supported by Stanford HAI, NSF RI \#2211258, ONR MURI N00014-22-1-2740, AFOSR YIP FA9550-23-1-0127, Amazon, Analog Devices, Bosch, JPMorgan Chase, Meta, and Salesforce.

{\small
\bibliographystyle{inc/ieee_fullname}
\bibliography{egbib}
}

\clearpage
\onecolumn

\begin{center}
{\Large \bf Supplementary for \model:\\Neuro-Symbolic Grounding of 3D Objects and Relations}
\end{center}

\vspace{1em}

\appendix

The appendix is organized as the following. In Appendix~\ref{sectapp:dsl}, we formally define the domain-specific language (DSL) used by \model. In Appendix~\ref{sectapp:experiments}, we provide dataset details, additional qualitative examples on both 3D-REC and 3D-QA tasks, and an additional experiment on scene complexity generalization, where we train models on scenes with a small number of objects but test on larger and more complex scenes.

\section{Domain-Specific Language}
\label{sectapp:dsl}

In this section, we summarize the value types (Table~\ref{tab:3dqa_dsl_types}) and function definitions (Table~\ref{tab:3dqa_dsl_operations}) of the domain-specific language used in our paper.
The \textit{scene} operation is parameter-free, while other functions take input \textit{object\_set}'s and output an \textit{object\_set}, represented as the object score vector. Query-type operations take input \textit{object\_set}'s and output answers of the target type, such as Boolean values (\eg, ``is there a chair?'') and concept names (\eg, ``what is the type of the object next to the table?''). Existence and counting-related operations involve a score threshold $t = 0.8$, which is a scalar hyperparameter. In our experiment, the threshold $t$ is chosen over a separate 3D-QA dataset based on scenes from the train set, instead of the test set. The \textit{query\_object}, \textit{query\_relation}, and  \textit{query\_t\_relation} operations are implemented through finding the object or relation label based on object score vectors.

\paragraph{Zero-shot transfer to 3D-QA.}
\model composes learned models on 3D-REC to build new 3D-QA operators in a zero-shot manner, requiring no additional training. The 3D-QA modules can be implemented by reusing the MLPs learned for object and relation classification from the 3D-REC task. Intuitively, let us consider \textit{query\_object} in 3D-QA, which takes an object as input and outputs its category. Since we have already learned classifiers for all categories (MLPs used in the \textit{filter} operation), \model directly reuses these modules to answer the question: it evaluates all MLP classifiers on the object feature and returns the category with the highest score.

\begin{table*}[hp!]
\centering\small
\renewcommand{\arraystretch}{1.45}
\setlength{\tabcolsep}{20pt}
\begin{tabular}{@{}lll@{}}
\toprule
Type  & Representation & Semantics \\ \midrule
\textit{object\_set} & Object score vectors.  & Set of objects selected from a scene. \\ \midrule
\textit{category}    & Concept names: {\it table, chair, piano, \etc.} & Object-level properties.           \\ \midrule
\textit{relation}    & Concept names: {\it near, left, behind, \etc.}  & (Binary) Relationships between two objects.           \\ \midrule
\textit{t\_relation}   & Concept names: {\it between, anchor-left \etc.}  & (Ternary) Relationships among three objects.           \\ \midrule
$^*$\textit{boolean} & Strings: {\it yes, no}. & Boolean values. \\ \midrule
$^*$\textit{integer} & Integers: {\it 0, 1, 2, \etc.} & Count of objects. \\ \midrule
\end{tabular}
\caption{Types in the \model domain-specific language. $^*$: Types that are only used in the 3D-QA task.}
\label{tab:3dqa_dsl_types}
\end{table*}

\begin{table*}[tp!]
\centering\small
\renewcommand{\arraystretch}{1.45}
\begin{tabular}{p{0.6\textwidth} p{0.35\textwidth}}
\toprule
Signature \& Implementation & Semantics \\
\midrule
\mycell{\textit{scene}() $\longrightarrow$ $y$: \textit{object\_set} \\
{\bf Implementation:}
$y_i = 0$, for all $i \in \{1,2,\cdots,M\}$}  & Return all objects in the 3D scene. \\
\midrule
\mycell{\textit{filter}($x$: \textit{object\_set}, $c$: \textit{category}) $\longrightarrow$ $y$: \textit{object\_set} \\
{\bf Implementation:}
$y_i = \min\left( x_i, \textit{prob}^\text{c}_i \right) = \min\left( x_i, \mathrm{MLP}^\text{c}\left( f^\text{obj}_i \right) \right)$} & Return all objects satisfying a concept \textit{c}. \\
\midrule
\mycell{\textit{relate}($x^t$: \textit{object\_set}, $x^r$: \textit{object\_set}, $\textit{rel}$: \textit{relation}) $\longrightarrow$ $y$: \textit{object\_set} \\
{\bf Implementation:}
$\textit{prob}^\text{rel}_{i, j} = \mathrm{MLP}^{\text{rel}}(f^\text{rel}_{i, j})$ \\
\qquad $y_i = \min\left(x^t_i, \sum_j \textit{sx}(x^r)_j \cdot \textit{prob}^\text{rel}_{i, j} \right)$} & Return all objects that satisfy the relationship \textit{rel} between the object sets.
\\ \midrule
\mycell{\textit{ternary\_relate}($x^t$: \textit{object\_set}, $x^\textit{r1}$: \textit{object\_set}, $x^\textit{r2}$: \textit{object\_set},\\\qquad\qquad\qquad $\textit{trel}$: \textit{t\_relation}) $\longrightarrow$ $y$: \textit{object\_set} \\
{\bf Implementation:}
$\textit{prob}^\text{trel}_{i, j, k} = \mathrm{MLP}^{\text{trel}}(f^\text{ternary}_{i, j, k})$ \\
\qquad $y_i = \min\left(x^t_i, \sum_j \sum_k \textit{sx}(x^{r1})_j \cdot \textit{sx}(x^{r2})_k \cdot \textit{prob}^\text{trel}_{i, j, k} \right)$} & Return all objects that satisfy the ternary relationship \textit{trel} between the object sets.
\\ \midrule
\mycell{\textit{anchor}(\textit{object\_set}, \textit{object\_set}) $\longrightarrow$ \textit{object\_set}\\
{\bf Implementation:} internally handled using {\it ternary\_relate}.
}
& Return all objects that satisfy the relationship anchored on the first object set.
\\ \midrule
\mycell{$^*$\textit{query\_exist}($x$: \textit{object\_set}) $\longrightarrow$ $y$: \textit{boolean} \\ {\bf Implementation:}
$y =
\begin{cases}
\textit{yes} &\text{if } \max_i(\sigmoid(x_i)) > t\\
\textit{no} &\text{if } \max_i(\sigmoid(x_i)) \leq t
\end{cases}
$} & Return Yes/No corresponding to existence of object in the object set. \\
\midrule
\mycell{$^*$\textit{query\_count}($x$: \textit{object\_set}) $\longrightarrow$ $y$: \textit{integer} \\ {\bf Implementation:} $y = \sum_i\mathbbm{1}\left[\sigmoid(x_i) > t\right]$} & Return count of objects in the object set. \\
\midrule
\mycell{$^*$\textit{query\_object}($x$: \textit{object\_set}) $\longrightarrow$ $c$: \textit{category} \\
{\bf Implementation:}
$c = {\arg\max}_{\text{c}} \left(\sum_i sx(x)_i \cdot \mathrm{MLP}^\text{c}\left( f^\text{obj}_i \right) \right)$} & Return type of object in the object set. \\
\midrule
\mycell{$^*$\textit{query\_relation}($x^t$ \textit{object\_set}, $x^r$: \textit{object\_set}) $\longrightarrow$ $\textit{rel}$: \textit{relation} \\
{\bf Implementation:} $\textit{rel} = {\arg\max}_\text{rel} \left(\sum_i \sum_j \textit{sx}(x^t)_i \cdot \textit{sx}(x^r)_j \cdot \mathrm{MLP}^{\text{rel}}(f^\text{rel}_{i, j}) \right)$} & Return relationship between the object sets. \\
\midrule
\mycell{$^*$\textit{query\_t\_relation}($x^t$:\textit{object\_set}, $x^{\textit{r1}}$: \textit{object\_set}, $x^{\textit{r2}}$: \textit{object\_set}) $\rightarrow$ $\textit{trel}$: \textit{t\_relation} \\
{\bf Implementation:} $\textit{trel} =$\\
\qquad ${\arg\max}_\text{trel} \left(\sum_i \sum_j \sum_k \textit{sx}(x^{t})_i \cdot \textit{sx}(x^\textit{r1})_j \cdot \textit{sx}(x^\textit{r2})_k \cdot
 \mathrm{MLP}^{\text{trel}}(f^\text{ternary}_{i, j, k}) \right)$
} & Return ternary relationship between the object sets.  \\
\bottomrule
\end{tabular}
\caption{Primitive functions defined in the \model domain-specific language. $^*$: Functions that are only used in the 3D-QA task. Here, $\textit{sx}(\cdot)$ is the Softmax function, $\sigma(\cdot)$ is the Sigmoid function, and $\mathbbm{1}[\cdot]$ is the indicator function which returns 1 when the expression inside the brackets evaluates to true, and 0 otherwise. }
\label{tab:3dqa_dsl_operations}
\end{table*}

\clearpage

\section{Experimental Details and Additional Results}
\label{sectapp:experiments}

In this section, we first present details for the datasets used in the main text. Then, we provide additional results on the scene complexity generalization task, where we train models on scenes with a small number of objects but test on larger and more complex scenes. Finally, we showcase additional qualitative examples for both the 3D-REC and 3D-QA tasks. %

\subsection{Dataset}

\paragraph{ReferIt3D datasets.}
For settings where \model was trained on the full ReferIt3D dataset, we used the exact SR3D training data for all networks, including $707$ scenes with object category annotations and $65,844$ query-answer pairs in total.

\paragraph{Data efficiency datasets.}
We generated data-efficient train sets with randomly sampled 0.5\% (329 examples), 1.5\% (987 examples), 2.5\% (1,646 examples), 5\% (3,292 examples), and 10\% (6,584 examples) of the train set, with the same full test set from SR3D used for evaluation. %

\paragraph{\textit{PAIRS} and \textit{SCENE} generalization datasets.}
We created two new datasets to test generalization ability. Both of the datasets are built on the SR3D train and test set.

The first dataset (PAIRS)  evaluates performance on unseen object-relation-object pairs. The referring expressions in the train set include the top 5 percent of object-relation-object pairs: \ie, the referred object category, relation type, and the reference object category (\eg, chair-closest-door). The test set contains the bottom 95 percent of object-relation-object pairs in the long-tailed distribution. The train set and test set consists of 16,200 examples and 10,520 examples respectively.

The second dataset (SCENE) evaluates performance on an unseen scene type. The train set includes train examples with all scene types aside from that of ``living room'', while the test set only contains examples in living rooms. The train set and test set consists of 57,125 examples and 1,320 examples respectively. %

\paragraph{3D-QA dataset.}
We manually created a small evaluation set of $50$ examples for the 3D-QA task, based on the test set of ReferIt3D \cite{achlioptas2020referit3d}. The input is a set of objects in the scene, $\mathcal{O} = \{O_1, ..., O_M\}$, and a question $\gQ$. In contrast to the 3D-REC task, where the output is the target object, the output for 3D-QA is an answer in text form (the vocabulary contains all categories, relations, Yes/No, and integers). The dataset consists of four main types of questions created from the following templates:

\vspace{4pt}
\noindent {\it Existence-typed questions:}
\begin{itemize}[noitemsep,topsep=4pt]
\item Is there a [\texttt{Object}] [\texttt{Relation}] [\texttt{Object}]?\qquad A: Yes/No
\item Is there a [\texttt{Object}] [\texttt{Relation}] [\texttt{Object}] and [\texttt{Object}]?\qquad A: Yes/No
\item Facing [\texttt{Object}], is there a  [\texttt{Object}] [\texttt{Relation}] [\texttt{Object}]?\qquad A: Yes/No
\end{itemize}

\vspace{4pt}
\noindent {\it Counting-typed questions:}
\begin{itemize}[noitemsep,topsep=4pt]
\item How many [\texttt{Object}] are in the scene?\qquad A: Integer
\item How many [\texttt{Object}] are [\texttt{Relation}] [\texttt{Object}]?\qquad A: Integer
\end{itemize}

\vspace{4pt}
\noindent {\it Object-typed questions:}
\begin{itemize}[noitemsep,topsep=4pt]
\item What is the item [\texttt{Relation}] [\texttt{Object}]?\qquad A: [\texttt{Object}]
\item What is the item [\texttt{Relation}] [\texttt{Object}] and [\texttt{Object}]?\qquad A: [\texttt{Object}]
\item Facing [\texttt{Object}], what is the item [\texttt{Relation}] [\texttt{Object}]?\qquad A: [\texttt{Object}]
\end{itemize}

\vspace{4pt}
\noindent {\it Relation-typed questions:}
\begin{itemize}[noitemsep,topsep=4pt]
\item What is the relationship between [\texttt{Object}] and [\texttt{Object}]?\qquad A: [\texttt{Relation}]
\item Facing [\texttt{Object}], what is the relationship between [\texttt{Object}] and [\texttt{Object}]?\qquad A: [\texttt{Relation}]
\end{itemize}

\clearpage

\begin{wraptable}{r}{0.5\textwidth}
\setlength{\tabcolsep}{12pt}
\begin{center}
\vspace{-0.5cm}
\begin{sc}
\begin{tabular}{lll}
    \toprule
     & Overall & View-dep. \\
    \midrule
    \model (Ours) & $\mathbf{0.627}$ & $\mathbf{0.620}$  \\
    MVT \cite{huang2022multi} & $0.405$ & $0.396$  \\
    SAT \cite{yang2021sat} & $0.444$ & $0.415$ \\
    TransRefer \cite{he2021transrefer3d} & $0.360$ & $0.344$ \\
    \bottomrule
  \end{tabular}
  \caption{Generalization results from sparse scenes to dense scenes.}
  \label{table:scene_sparsity}
\end{sc}
\end{center}
\vspace{-1em}
\end{wraptable}

\subsection{Scene Complexity Generalization}
For all experiment results reported in the main text, \model was trained on examples with only $10$ objects given in the scene, and evaluated on the full test set with up to $88$ objects in the scene. \model is able to show this scene complexity generalization, as it does not need the full scene point cloud as its input and instead only explicitly models a given object set and relations between specified objects. This improves training efficiency, reduces the need for annotated 3D objects, which are expensive to acquire in 3D domains, and enables generalization to more cluttered scenes.

We show that baselines methods cannot generalize as \model does, and yields significantly decreased performance when trained on $10$ objects per scene and evaluated on more complex scenes.
In Table~\ref{table:scene_sparsity}, we see that \model outperforms prior works by a large margin in this setting. We did not test BUTD-DETR, because BUTD-DETR explicitly encodes the full 3D scene as input, with all objects given in train and test, and hence does not directly apply to this partial scene setup.

\subsection{NR3D Results}

\begin{wraptable}{r}{0.5\textwidth}
\setlength{\tabcolsep}{12pt}
\begin{center}
\vspace{-0.5cm}
\begin{sc}
\begin{tabular}{lll}
    \toprule
     & Overall & View-dep. \\
    \midrule
    \model (Ours) & $\mathbf{0.526}$ & $\mathbf{0.432}$  \\
    BUTD-DETR \cite{jain2022bottom} & $0.382$ & $0.331$ \\
    MVT \cite{huang2022multi} & $0.430$ & $0.324$  \\
    SAT \cite{yang2021sat} & $0.329$ & $0.248$ \\
    TransRefer \cite{he2021transrefer3d} & $0.360$ & $0.286$ \\
    \bottomrule
  \end{tabular}
  \caption{Results on a constrained version of NR3D from the ReferIt3D datasets.}
  \label{table:nr3d}
\end{sc}
\end{center}
\vspace{-1em}
\end{wraptable}

We report results on NR3D, the natural language variant of ReferIt3D. While \model does work on natural language, as Codex can parse NR3D input into programs, Codex parsing yields $91$ distinct function modules and $5892$ concepts, resulting in a separate long-tailed problem. Hence, we ran additional experiments on a subset of NR3D, by restricting utterances to those that parse to the same set of functions and concepts in SR3D, which yields $3659$ train examples and  $1041$ test examples.

We train \model as well as top-performing baselines, and see that \model significantly outperforms prior work (Table~\ref{table:nr3d}). This suggests that \model can learn from natural language data in a data-efficient way. Examples from this NR3D subset include noisy natural language such as ``The picture above the bed with the laptop on it.'' and ``The monitor that you would class as in the middle of the other two''; both exhibit noisy natural language, with more complex underlying programs than the SR3D training set.

\subsection{Qualitative Examples}
\begin{figure}[tp!]
  \centering
    \includegraphics[width=\linewidth]{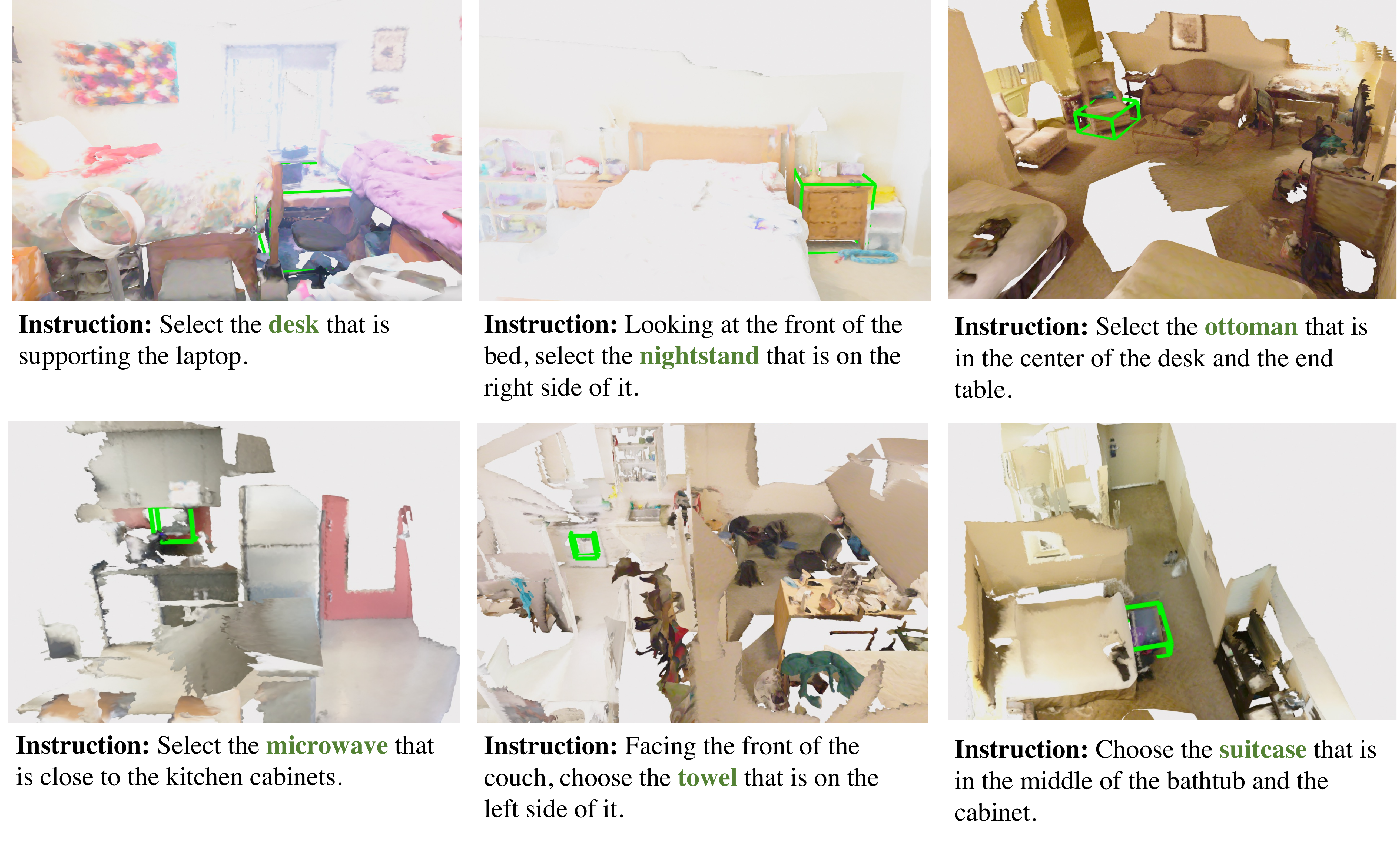}
  \caption{Additional examples of the input language instruction, scene, and the target output in the ReferIt3D 3D-REC task.}
\label{fig:appendix_rec}
\end{figure}

\begin{figure*}[tp!]
  \centering
    \includegraphics[width=1.0\linewidth]{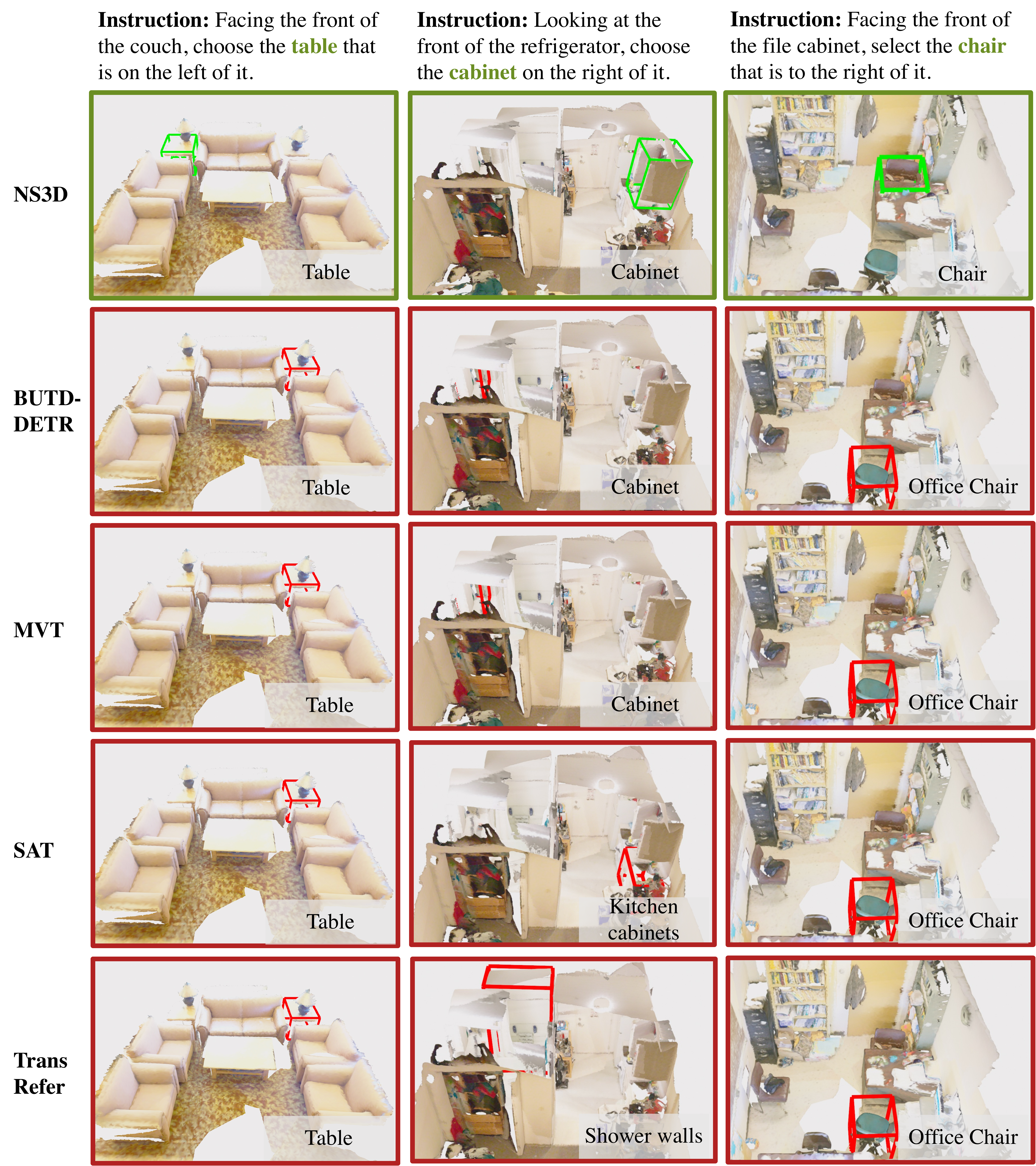}
  \caption{Comparison between \model and baselines on the 3D-REC task, with the green outline indicating correct selection and red outline indicating incorrect selection.}
\label{fig:appendix_rec_comp}
\end{figure*}

In Figure~\ref{fig:appendix_rec}, we show additional examples of the ReferIt3D 3D-REC task in SR3D, with examples of binary and ternary relations. In Figure~\ref{fig:appendix_rec_comp}, we present comparisons of \model against baselines on view-dependent examples, with the green outline indicating correct selection and red outline indicating incorrect selection. We see that \model is able to outperform prior work in disambiguating the target referred object.

\begin{figure*}[tp!]
  \centering
    \includegraphics[width=0.75\linewidth]{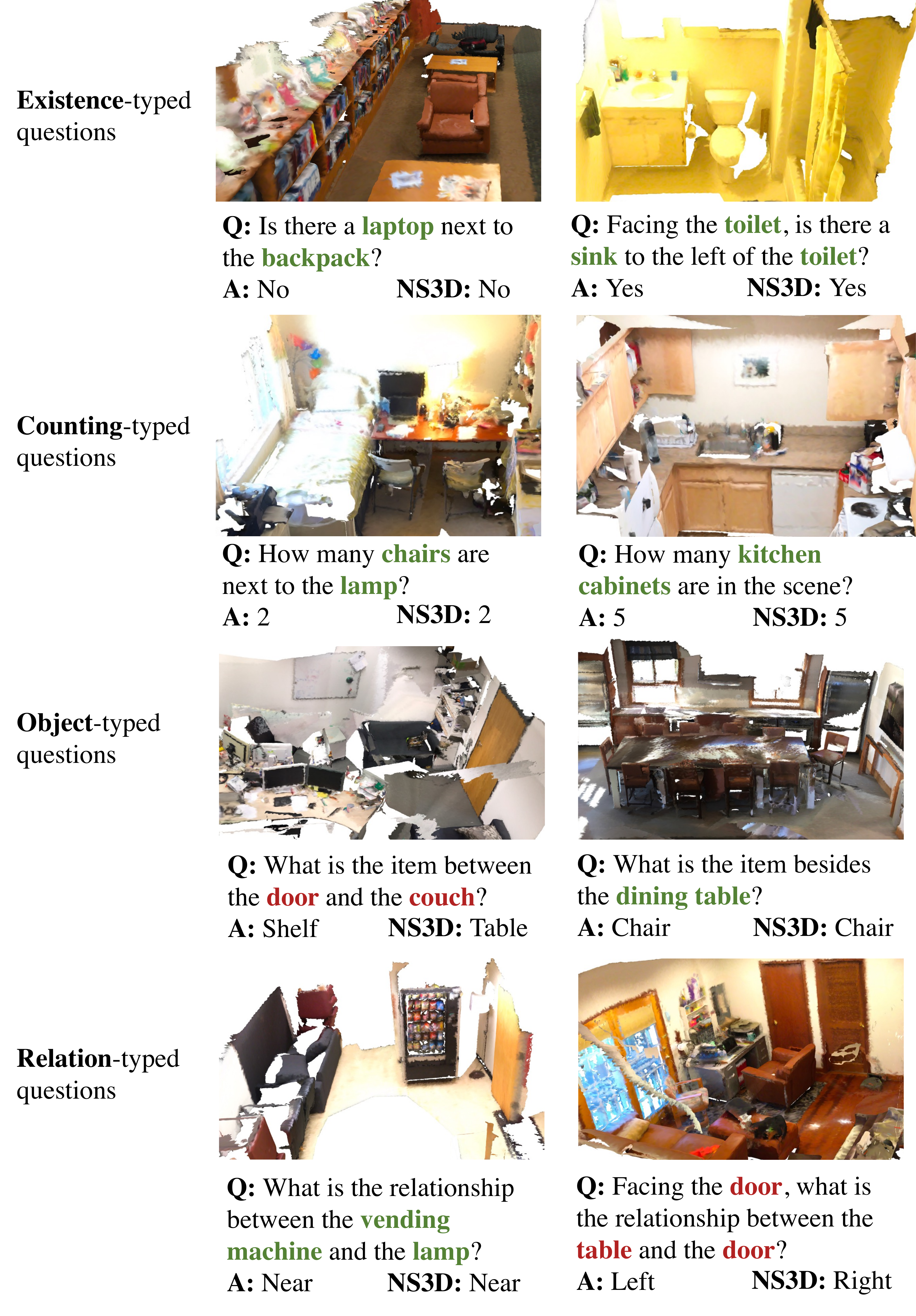}
  \caption{Qualitative examples of \model on the 3D-QA task. Examples of success cases are marked in green, while failure cases are marked in red.}
\label{fig:appendix_qa}
\end{figure*}

In Figure~\ref{fig:appendix_qa}, we present additional qualitative examples of \model on the 3D-QA task. We see examples of success cases in green and failure cases in red for \model in the zero-shot transfer setting.

\end{document}